\documentclass{article} 
\usepackage{iclr2025_conference,times}

\usepackage{hyperref}
\usepackage{url}
\usepackage{algorithm,algorithmicx,algpseudocode}
\usepackage{amsmath}
\usepackage{subcaption}
\usepackage{amsthm}
\usepackage{caption}
\usepackage{dsfont}
\usepackage{mathtools}
\usepackage{booktabs}
\usepackage{array}
\usepackage{siunitx}
\usepackage{makecell}
\usepackage{amssymb}
\usepackage{multirow}
\usepackage[toc,page,header]{appendix}
\usepackage{minitoc}
\usepackage{soul}
\usepackage{tocloft}
\usepackage{amsmath}
\usepackage{changepage}
\usepackage{textcomp} 
\usepackage{wrapfig}
\usepackage{tcolorbox}
\usepackage{enumitem}
\usepackage{url}
\usepackage{adjustbox}
\usepackage{pifont}
\usepackage{tabularx}
\usepackage{amsmath}
\usepackage{bm}
\usepackage{amsfonts}
\usepackage{bibentry}
\usepackage{gensymb}
\usepackage[capitalize]{cleveref}
\usepackage{fontawesome}
\usepackage[T1]{fontenc}
\crefname{equation}{Eq.}{Eqs.}
\Crefname{equation}{Equation}{Equations}
\crefname{section}{\S}{\S\S}
\Crefname{section}{Section}{Sections}
\Crefname{table}{Table}{Tables}
\crefname{table}{Tab.}{Tabs.}
\crefname{appendix}{App.}{Apps.}
\Crefname{appendix}{Appendix}{Appendices}

\tcbuselibrary{listingsutf8}

\newtheorem{definition}{Definition}

\newcommand\numberthis{\addtocounter{equation}{1}\tag{\theequation}}

\newtcbox{\hlight}[1][]{on line, colback=green!30, colframe=green!30, boxsep=0pt, left=1pt, right=1pt, top=1pt, bottom=1pt}

\newcolumntype{M}[1]{>{\centering\arraybackslash}m{#1}}
\newcolumntype{L}[1]{>{\raggedright\arraybackslash}m{#1}}
\newcommand{\mypara}[1]{\noindent\textbf{#1\quad}}

\def\ourmethod{\textsc{CoInD}}
\definecolor{skyblue}{RGB}{232, 243, 255}
\definecolor{darkpink}{RGB}{255,20,147}
\definecolor{fireenginered}{rgb}{0.81, 0.09, 0.13}
\definecolor{lightgreen}{RGB}{144,238,144}

\newcommand{\jsd}{\ensuremath{\text{JSD}}}

\newcommand{\Cspace}{\ensuremath{\mathcal{C}}}
\newcommand{\Ctrain}{\ensuremath{\Cspace_{\text{train}}}}

\newcommand{\cs}{\ensuremath{\text{CS}}}

\newcommand{\ind}{\perp\!\!\!\perp} 
\newcommand{\notind}{\not\!\perp\!\!\!\perp} 

\def\eqref#1{equation~\ref{#1}}

\def\1{\bm{1}}

\DeclareMathAlphabet{\mathsfit}{\encodingdefault}{\sfdefault}{m}{sl}
\SetMathAlphabet{\mathsfit}{bold}{\encodingdefault}{\sfdefault}{bx}{n}

\newcommand{\X}{\ensuremath{\bm{X}}} 

\newcommand{\pdata}{p_{\text{data}}}

\newcommand{\E}{\mathbb{E}}

\newcommand{\lci}{\mathcal{L}_{\text{CI}}}
\newcommand{\lscore}{\mathcal{L}_{\text{score}}}
\newcommand{\score}{\nabla_{\mathbf{X}} \log}

\newcommand{\redhighlight}[1]{{\sethlcolor{red!10}\hl{#1}}}
\newcommand{\greenhighlight}[1]{{\sethlcolor{green!10}\hl{#1}}}

\iclrfinalcopy
\title{Compositional World Knowledge leads to High Utility Synthetic data}

\author{Sachit Gaudi \quad Gautam Sreekumar \quad Vishnu Naresh Boddeti\\
Michigan State University\\
\texttt{\{gaudisac,sreekum1,vishnu\}@msu.edu}
}

\begin{document}

\doparttoc 
\faketableofcontents 

\maketitle
\begin{abstract}
Machine learning systems struggle with robustness, under subpopulation shifts. This problem becomes especially pronounced in scenarios where only a subset of attribute combinations is observed during training—a severe form of subpopulation shift, referred as \textit{compositional shift}. To address this problem, we ask the following question: Can we improve the robustness by training on synthetic data, spanning all possible attribute combinations? We first show that training of conditional diffusion models on limited data lead to incorrect underlying distribution. Therefore, synthetic data sampled from such models will result in unfaithful samples and does not lead to improve performance of downstream machine learning systems. To address this problem, we propose \ourmethod{} to reflect the compositional nature of the world by enforcing conditional independence through minimizing Fisher's divergence between joint and marginal distributions. We demonstrate that synthetic data generated by \ourmethod{} is faithful and this translates to state-of-the-art worst-group accuracy on \textit{compositional shift} tasks on CelebA. Our code is available at \color{darkpink}{\href{https://github.com/sachit3022/compositional-generation/}{https://github.com/sachit3022/compositional-generation/} }

\end{abstract}

\begin{figure}[ht]
\centering
\includegraphics[width=0.88\linewidth]{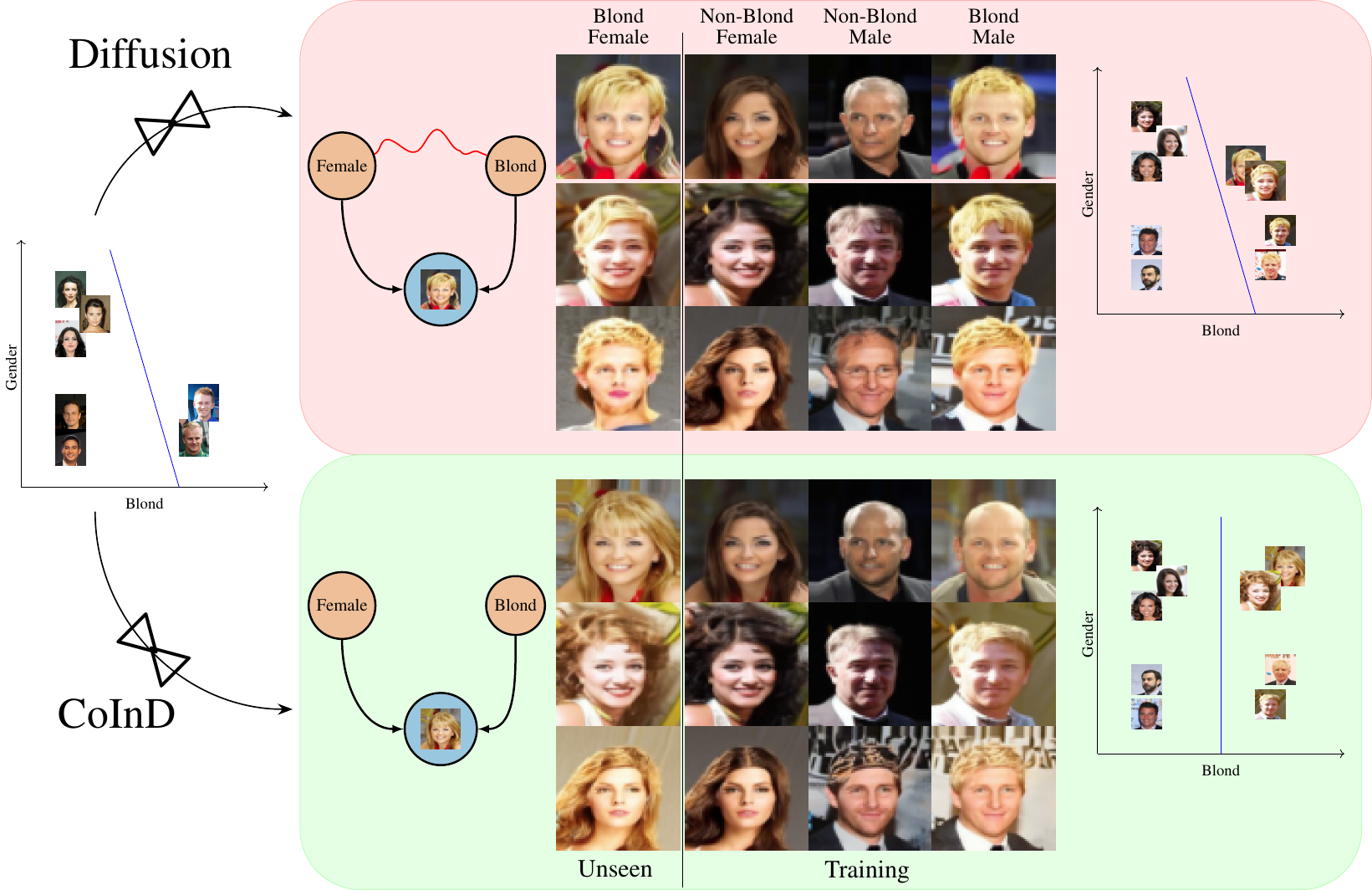}
    \caption{
Training data comprises of all combinations except female blonds. Classifier trained to predict \textit{blond} relies on information of \textit{gender} due to their association in training data. \redhighlight{Standard diffusion model trained on this data also learns this association leading to incorrect distribution, resulting in unfaithful generation of unseen combinations (female blondes).Consequently, synthetic data from this model fails to improve downstream classifier performance.} \greenhighlight{In contrast, \ourmethod{} leverages compositional world knowledge to learn the true distribution, facilitating accurate generation of unseen combinations. This leads to a robust classifier when trained on synthetic data from \ourmethod{}.} {\label{fig:teaser}}}
\end{figure}

\section{Introduction\label{sec:introduction}}

Consider CelebA~\cite{liu2015faceattributes}  dataset with 40 binary attributes would require over 1 trillion samples to span all combinations. Collecting such massive data for all attribute combinations is infeasible; Machine learning systems trained on subset of combinations will suffer from the problem of \textit{compositional shift}~\citep{mahajan2024compositional}. Consider a simplified base case of \textit{compositional shift}, where \textit{female blonds} are not observed during training but all the other combinations of gender, blond are observed as depicted in~\cref{fig:teaser}. 

Humans can easily synthesize and compose attributes such as blond and gender, and therefore able to imagine \textit{female blonds} from just observing \textit{male blonds}, \textit{female non-blonds}, and \textit{male non-blonds}~\citep{baroni2023human}. In this work, we want to learn generative models that have the capability to imagine. This imagination can be distilled to generate synthetic samples spanning all combinations. The success of the task depends on the models ability to faithfully generate unseen compositions. Standard diffusion models~\cite{NEURIPS2020_4c5bcfec} are trained with an optimization designed to maximize likelihood, which results in memorization of training data rather than true generalization~\citep{kamb2024analytictheorycreativityconvolutional}. Diffusion models either fail to respect the gender, or learn incorrect interpolations between blond and gender attributes. The samples in~\cref{fig:teaser}(red box) reveal these limitations. 

The difference between humans and diffusion models is that we understand the compositional nature of the world, allowing us to create complex composites from a set of primitive components~\cite{nye2020learning}. However, diffusion models learn the associations from the data. In fact, we verify that diffusion models trained on limited data violate conditional independence, which is an important assumption in compositionality~\cite{nie2021controllable,liu2021learning}. We point out that this violation stems from the incorrect objective of diffusion models under limited data. Therefore, we propose \ourmethod{} to incorporate compositional world knowledge into diffusion model training by minimizing the fisher's divergence between conditional joint and product of conditional marginals, in addition to maximizing likelihood on the observed compositions. 

\ourmethod{} exhibits compositionality, effectively learning the true underlying data distribution. This results in the faithful generation of previously unseen attribute combinations, as clearly illustrated in the~\cref{fig:teaser}(green box). Notably, classifiers trained on this higher-quality synthetic data exhibit enhanced robustness and generalization capabilities, compared to standard diffusion models. These classifiers achieve SoTA results, significantly outperforming established baselines in subpopulation shift literature. Moreover, \ourmethod{} offers a remarkably simple implementation, requiring only a few additional lines of code to standard diffusion model training.

\section{Compositional World \label{sec:problem-def}}
\begin{wrapfigure}{r}{0.25\textwidth}
    \centering
    \vspace{-1.0cm}
    \includegraphics[width=0.24\textwidth]{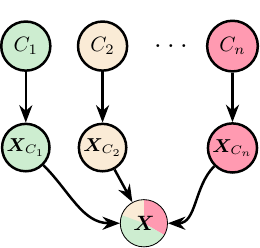}
    \caption{\textbf{Compositional world model}. $\X$ is generated by varying $C_1, C_2, \dots, C_n$  independently in the underlying data generating process.\label{fig:causal-graph-inf}}
    \vspace{-0.4cm}
\end{wrapfigure}

We study the model of learning compositional functions from limited data. As compositionality is a property of data generating process~\citep{wiedemer2024compositional}. We consider the case, where the samples are generated by independently varying factors, and have access to the labels of these factors. Similar assumptions were discussed in \citep{wiedemer2024compositional,ahmed2020systematic}. In this section, we formally define the assumption in the data generation process.

\mypara{Notations.}We use bold lowercase and uppercase characters to denote vectors~(e.g., $\bm{a}$) and matrices~(e.g., $\bm{A}$) respectively. Random variables are denoted by uppercase Latin characters (e.g., $X$). The distribution of a random variable $X$ is denoted as $p(X)$, or as $p_{\bm{\theta}}(X)$ if the distribution is parameterized by a vector $\bm{\theta}$. With a slight abuse of notation, we refer to $p_{\bm{\theta}}(\X\mid C_i)$ as marginal, and joint as $p_{\bm{\theta}}(\X\mid C)$.

\mypara{Data Generation Process.}The data generation process consists of observed data $\X$~(e.g., images) and its attribute variables $C_1, C_2, \dots, C_n$~(e.g., color, digit, etc.). To have explicit control over these attributes during generation, they should vary independently of each other. In this work, we limit our study to only those causal graphs in which the attributes are not causally related and can hence vary independently, as shown in \cref{fig:causal-graph-inf}. Each $C_i$ assumes values from a set $\Cspace_i$ and their Cartesian product $\Cspace = \Cspace_1 \times \dots \times \Cspace_n$ is referred to as the \textit{attribute space}. Each attribute $C_i$ generates its own observed component $\X_{C_i} = f_{C_i}(C_i)$, which together with unobserved exogenous variables $\bm{U}_{\X}$ form the composite observed data $\X = f(\X_{C_1},\dots, \X_{C_i}, \bm{U}_{\X})$ (see \cref{fig:causal-graph-inf}). We do not restrict $f$ much except that it should not obfuscate individual observed components in $\X$~\citep{wiedemer2024compositional}. A simple example of $f$ is the concatenation function. We also assume that all $f_{C_i}$ are invertible and therefore it is possible to estimate $C_1,\dots, C_n$ from $\X$. These assumptions together ensure that $C_1, \dots, C_n$ are mutually independent given $\X$ despite being seemingly d-connected.

The attribute space in our problem statement has the following properties. \textbf{(1)}~The attribute space observed during training $\Ctrain$ covers $\Cspace$ in the following sense:
\begin{definition}[Support Cover]
 Let $\Cspace = \Cspace_1\times\dots\times\Cspace_n$ be the Cartesian product of $n$ finite sets $\Cspace_1, \dots, \Cspace_n$. Consider a subset $\Ctrain\subset\Cspace$. Let $\Ctrain = \{(c_{1j},\dots,c_{nj}): c_{ij}\in\Cspace_i, 1 \leq i \leq n, 1 \leq j \leq m\}$ and $\tilde{\Cspace}_i = \{c_{ij} : 1\leq j \leq m\}$ for $1\leq i \leq n$. The Cartesian product of these sets is $\tilde{\Cspace}_{\text{train}} = \tilde{\Cspace}_1\times\dots\times\tilde{\Cspace}_n$. We say $\Ctrain$ covers $\Cspace$ iff $\Cspace = \tilde{\Cspace}_{\text{train}}$.
\end{definition}
Informally, this assumption implies that every possible value that $C_i$ can assume is present in the training set, and open-set attribute compositions do not fall under this definition. \textbf{(2)}~For every ordered tuple $c\in\Ctrain$, there is another $c'\in\Ctrain$ such that $c$ and $c'$ differ on only one attribute.

In the example discussed in~\cref{sec:introduction}, $C_1$ represents the blond hair and $C_2$ encodes gender. Blond and gender independently generate an image, ($\X$), while both blond and gender can be reversely inferred from the image. We observe all attribute combinations except $(C_1,C_2)=(1,0)$ (where $C_1=1$ denotes blond and $C_2=0$ indicates female), consistent with the assumptions outlined earlier.

\mypara{Preliminaries on Score-based Models}In this work, we train conditional score-based models~\citep{song2021scorebased} using classifier-free guidance~\citep{ho2022classifier} to generate data corresponding by composing attributes. Score-based models learn the score of the observed data distributions $p_{\text{train}}(\X)$ and $p_{\text{train}}(\X\mid C)$ through score matching~\citep{hyvarinen2005estimation}. Once the score of a distribution is learned, samples can be generated using Langevin dynamics. \citet{liu2023compositional} proposed the following modifications during sampling to enable compositionality, assuming that the model learns the conditional independence relations from the data-generation process.

\noindent\textbf{Compositional Sampling:} $C_1 = c_1\land C_2=c_2$ generates data where attributes $C_1$ and $C_2$ takes values $c_1$ and $c_2$ respectively. Since, $p_{\bm{\theta}}(C_1 \land C_2 \mid \X) = p_{\bm{\theta}}(C_1 \mid \X)p_{\bm{\theta}}(C_2 \mid \X)$ samples are generated for the composition $C_1\land C_2$ by sampling from the following score:
\begin{align}
    \nabla_{\X} \log p_{\bm{\theta}}(\X \mid C_1 \land C_2 ) = \nabla_{\X} \log p_{\bm{\theta}}(\X \mid C_1 ) + \nabla_{\X} \log p_{\bm{\theta}}(\X \mid C_2) - \nabla_{\X} \log p_{\bm{\theta}}(\X) \label{eq:and-score}
\end{align}
This formulation gives can me modified for additional flexibility of controlling for attributes, where $\gamma$ controls the strength of attribute $C_1$ w.r.t $C_2$,  $\nabla_{\X} \log p_{\bm{\theta}}(\X \mid C_1 \land C_2 )$ can be written as:
\begin{align}
    \gamma \nabla_{\X} \log p_{\bm{\theta}}(\X \mid C_1 ) + \nabla_{\X} \log p_{\bm{\theta}}(\X \mid C_2) - \gamma\nabla_{\X} \log p_{\bm{\theta}}(\X) \label{eq:and-score-2}
\end{align}

\section{Diffusion models do not respect the  compositional world\label{sec:case-study}}

To achieve compositionality, diffusion models must preserve the properties of the underlying data distribution. As established in \cref{sec:problem-def}, a critical property is conditional mutual independence (CI). Additionally, this property is used to sample data using compositional sampling.  \begin{wraptable}{R}{4.5cm}  
\begin{tabular}{lcc}
\toprule
Support & \jsd{} $\downarrow$  & \makecell{ CS$\uparrow$}\\
\midrule
Full & 0.37 & 96.58 \\
Partial & 0.62 & 41.80   \\
\bottomrule
\end{tabular}
\caption{Jensen-Shannon divergence and Conformity Scores for CelebA dataset.}
\vspace{-0.5cm}
\label{tab:cs-results}
\end{wraptable} 

We empirically verify whether diffusion models satisfy CI through two experimental configurations: \textbf{(1)~Full support}, where the model observes all ordered pairs in the concept space $\mathcal{C}$, and \textbf{(2)~Partial support}, limited to a training subset $\mathcal{C}_{\text{train}} \subset \mathcal{C}$ (visualized in \cref{fig:teaser}).

We measure CI violation as the disparity between the conditional joint distribution $p_{\bm\theta}(C\mid \X)$ and the product of conditional marginal distributions $\prod_{i}^n p_{\bm\theta}(C_i \mid \X)$ learned by the implicit classifier of diffusion model using Jensen-Shannon divergence as,
\begin{equation}\label{eq:wasser}
\jsd = \E_{C, \X \sim \pdata}\left[ D_{\text{JS}}\left( p_{\bm{\theta}}(C \mid \X) \mid\mid \prod_i^n p_{\bm{\theta}}(C_i \mid \X) \right) \right]
\end{equation}

where $D_{\text{JS}}$ is the Jensen-Shannon divergence and $p_{\bm{\theta}}$ is obtained by following \citep{Li_2023_ICCV} and evaluating the implicit classifier learned by the diffusion model. More details can be found in \cref{sec:jsd}.

A positive JSD value suggests that the model fails to adhere to the CI relation. As shown in~\cref{tab:cs-results}, models trained on partially observed attribute compositions (partial support) exhibit substantially higher JSD values than those trained with full composition coverage. This result demonstrates that diffusion models directly propagate spurious dependencies present in the training data rather than learning conditionally independent attributes.

Violation in conditional independence originates from the standard training objective of diffusion models that maximize the likelihood of conditional generation. Under perfect loss, for every observed composition ($C \in \mathcal{C}_{train}$), the model accurately learns $p_{\text{train}}(\X\mid C)$, i.e., $p_{\bm\theta}(\X\mid C) \approx p_{\text{train}}(\X\mid C) = p(\X\mid C)$, However, learn incorrect marginals, $p_{\bm\theta}(\X\mid C_i) \approx p_{\text{train}}(\X\mid C_i) \ne p(\X\mid C_i)$. Refer to \cref{appsubsec:incorrect-marginals} for complete proof. Informally, $p_{\text{train}}(\X\mid C_1=1)$, consists of only blond males. While, the underlying $p(\X\mid C_1=1)$ contains male, and females blonds. These incorrect marginals lead to violation in CI.

This violation leads to unfaithful generation, evident in lower conformity scores (CS) for partial-support models versus full-support (~\cref{tab:cs-results}). CS quantifies alignment between generated attributes (inferred via $\phi_{C_i}$) and conditioning attributes $(C)$ (details: \cref{cs-score}). Samples generated under partial support (highlighted in red, \cref{fig:teaser}) exhibit gender leakage —such as beards— when synthesizing unseen ``blond females". Based on these observations, we propose \ourmethod{} to train diffusion models that explicitly enforce the conditional independence to encourage compositionality.

\section{\ourmethod{}: Enforcing Conditional Independence Enables Compositionality\label{sec:method}}

In the previous section, we observed that diffusion models violate conditional independence (CI) by learning incorrect marginals. To remedy this, \ourmethod{} uses a training objective that explicitly enforces the causal factorization to ensure that the trained diffusion models obey CI. Applying the assumption of $C_1\ind\dots \ind C_n\mid \X$ mentioned in \cref{sec:problem-def}, we have $p(\X \mid C) = \frac{p(\X)}{p(C)}\prod_i^n\frac{p(\X\mid C_i)p(C_i)}{p(\X)}$. Note that the invariant $p(\X \mid C)$ is now expressed as the product of marginals employed for sampling. Therefore, training the diffusion model by maximizing this conditional likelihood is naturally more suited for learning accurate marginals for the attributes. We minimize the distance between the true conditional likelihood and the learned conditional likelihood as,
\begin{align}
    \mathcal{L}_{\text{comp}} = \mathcal{W}_2\left( p(\X \mid C), \frac{p_{\bm\theta}(\X)}{p_{\bm\theta}(C)} \prod_i \frac{p_{\bm\theta}(\X \mid C_i)p_{\bm\theta}(C_i)}{p_{\bm\theta}(\X)} \right) \label{eq:linv-original}
\end{align}
leveraging the Wasserstein distance upper bound via Fisher divergence~\citep{kwon2022scorebased}, we derive the following inequality:
\begin{equation}
    \mathcal{L}_{\text{comp}} \leq K_1\sqrt{\mathcal{L}_{\text{score}}} + K_2\sqrt{\mathcal{L}_{\text{CI}}}
    \label{eq:lcomp}
\end{equation}
for constants $K_1, K_2 > 0$. A complete derivation of this bound is provided in \cref{sec:proof}.

\textbf{Distribution matching:} 
\begin{align}
    \mathcal{L}_{\text{score}} = \E_{p({\X},C)}\lVert \nabla_{\X} \log p_{\bm{\theta}}(\X \mid C) - \nabla_{\X}\log p(\X \mid C)\rVert_2^2
   \label{eq:score-matching}
\end{align}
\textbf{Conditional Independence:}
\begin{align}
   \lci = \E\| \score p_\theta(\mathbf{X}\mid C) - \score p_\theta(\mathbf{X}) - \sum_i \left[\score p_\theta(\mathbf{X}\mid C_i) - \score p_\theta(\mathbf{X})\right]\|_2^2 \label{eq:cond-ind-score}
\end{align}

\textbf{Practical Implementation.} A computational burden presented by $\lci$ in \cref{eq:cond-ind-score} is that the required number of model evaluations increases linearly with the number of attributes. To mitigate this burden, we approximate the mutual conditional independence with pairwise conditional independence~\citep{hammond2006essential}. Thus, the modified $\lci{}$ becomes,

\begin{align*}
    \lci = \E_{p(\X,C)}\E_{j,k}\lVert\nabla_{\X} \log p_\theta(\X\mid C_j,C_k) - \nabla_{\X}\log  p_\theta(\X \mid C_j) -  \nabla_{\X}\log  p_\theta(\X \mid C_k)+\nabla_{\X} \log  p_\theta(\X)\rVert_2^2 \label{eq:lci_0}
\end{align*}
The weighted sum of the square of the terms in~\cref{eq:lcomp} has shown stability. Therefore, \ourmethod{}'s training objective:
\begin{equation}\label{eq:final_eq}
    \mathcal{L}_{\text{final}} = \mathcal{L}_{\text{score}} + \lambda \lci
\end{equation}
where $\lambda$ is the hyper-parameter that controls the strength of conditional independence. The reduction to the practical version of the upper bound (\cref{eq:lcomp}) is discussed in extensively in \cref{sec:practical}. For guidance on selecting hyper-parameters in a principled manner, please refer to \cref{sec:hyperparameter}. Finally, our proposed approach can be implemented with just a few lines of code, as outlined in \cref{alg:training}.

\section{Results}
We perform experiments on the CelebA dataset, downsampled to $64\times 64$. pixels. As a remainder, diffusion model was trained on all compositions except for ``blonde females."  We followed similar settings for both the standard diffusion model and \ourmethod{}($\lambda=100$). For detailed experimental settings, please refer to Appendix Section~\cref{appsec:experiment-details}

\subsection{Compositionality leads to high utility synthetic data}

\begin{figure}[H]
\vspace*{-0.3cm}
\begin{minipage}[c]{0.68\textwidth}
    \includegraphics[width=\linewidth]{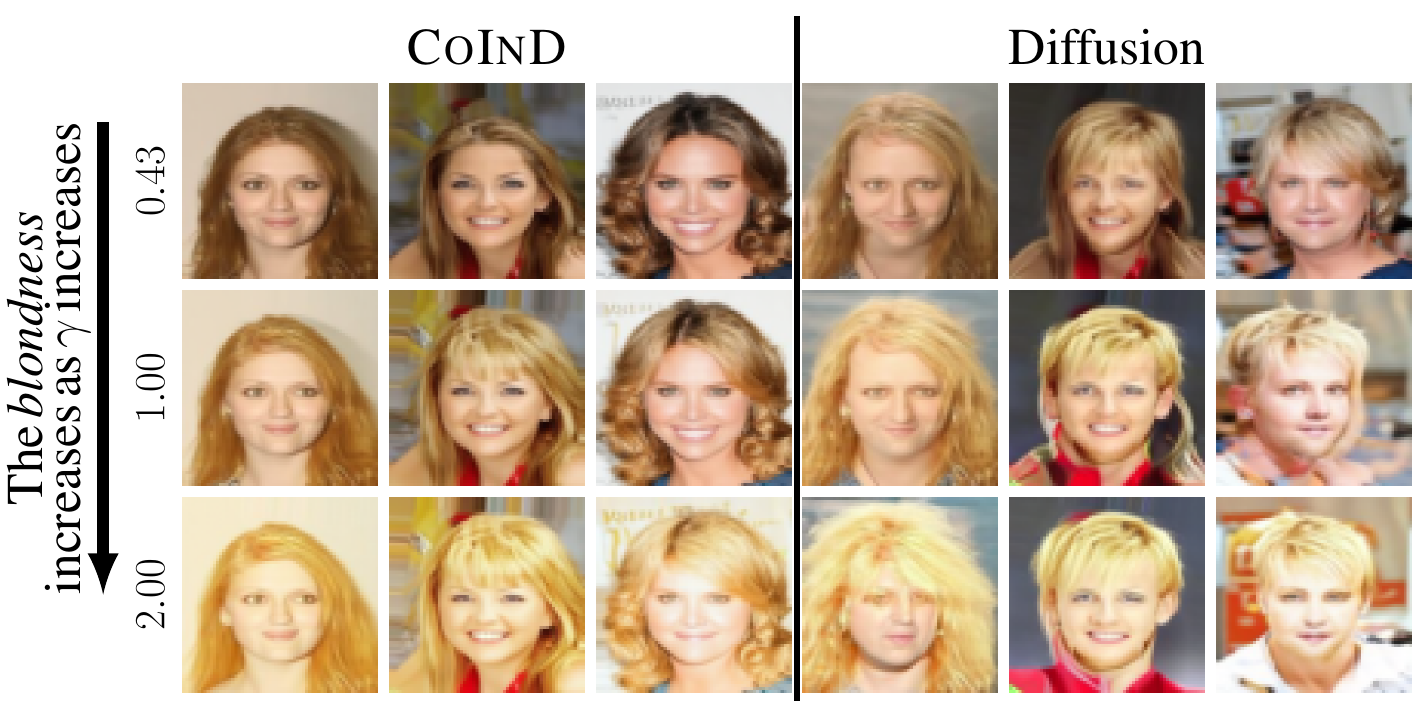}
\end{minipage}%
\vspace*{-0.2cm}
\hspace{0.2em}
\begin{minipage}[c]{0.28\textwidth}
    \caption{\ourmethod{} enables precise control~(\cref{eq:and-score-2}) over blondness, while preserving gender attributes (left). Standard diffusion models exhibit gender bias by conflating blondness with specific genders due to training data correlations (right).\label{fig:img_control}}
\end{minipage}
\vspace*{-0.2cm}
\end{figure}

\begin{wraptable}{R}{5.5cm}  
\vspace{-0.5cm}
\begin{tabular}{lccc}
\toprule
Method & \jsd{} $\downarrow$  &    FID $\downarrow$ & \makecell{ CS$\uparrow$}\\
\midrule
Diffusion & 0.62   & 41.80 & 64.77 \\
\ourmethod{} &  \textbf{0.16}  & \textbf{21.64} & \textbf{81.05 }\\
\bottomrule
\end{tabular}
\caption{JSD, CS, FID for CelebA dataset.}
\vspace{-0.3cm}
\label{tab:faith-results}
\end{wraptable} 

Our analysis, as shown in \cref{tab:faith-results}, demonstrates that  \ourmethod{} significantly reduces violations of conditional independence, resulting in more faithful data generation. \cref{fig:teaser}(highlighted in green) illustrates this improvement, showcasing  \ourmethod{}'s ability to accurately synthesize previously unseen ``blonde females" compositions. \ourmethod{} adapts hairstyles from female celebrities and blends blonde shades from male counterparts, producing photorealistic blonde female figures - despite not being explicitly trained on these examples.

\subsection{Synthetic Data Sampled from \ourmethod{} Yields Robust Classifier \label{sec:experiments}}

Building on the demonstrated capability of \ourmethod{} to generate high-utility synthetic data, this section addresses the following question: Can leveraging this data enhance classifier robustness against compositional shifts?

\mypara{Setup} We generate 20,000 samples of synthetic data from the trained diffusion model by uniformly sampling~(\cref{eq:and-score-2}) all possible compositions, including ``blonde females." We then train a ResNet-18 classifier on this synthetic data to predict the ``blonde" attribute. To evaluate performance, we measure the test accuracy, balanced test accuracy, and Worst Group Accuracy (WGA) on the downsampled ($64\times 64$) test set. We compare these results to the baselines borrowed from  ~\citep{mahajan2024compositional}.
 
\begin{table}
\centering
\begin{tabular}{llccc}
\toprule
\textbf{Data} & \textbf{Method} & \textbf{Test Acc.} & \textbf{Balanced Acc.} & \textbf{Worst group Acc.} \\
\midrule
\multirow{5}{*}{Real}&ERM & 87.0 & 59.3 & 4.0 \\
&G-DRO~\citep{sagawa2019distributionally} & 91.7 & \textbf{86.3} & 71.7 \\
&LC~\citep{liu2212avoiding} & 88.3 & 70.7 & 21.0 \\
&sLA~\citep{tsirigotis2024group} & 88.3 & 71.0 & 21.3 \\
&CRM~\citep{mahajan2024compositional} & \textbf{93.0} & 85.7 & 73.3 \\
\midrule
\multirow{2}{*}{Synthetic}& Diffusion & 90.9	&80.4&	64.0 \\
&CoInD & 80.6 &	80.3 & \textbf{76.9} \\
\bottomrule
\end{tabular}
\caption{\textbf{Classifier trained with synthetic data generated from \ourmethod{} achieves better Worst group Acc.} for classifying attribute ``blond'' attribute of CelebA dataset.}
\vspace{-0.5em}
\label{tab:method_comparison}
\end{table}

\mypara{Discussion} \ourmethod{} exhibits $\approx 5\%$ higher WGA compared to the baselines(\cref{tab:method_comparison}), which makes it robust to compositional shift. However, other models exhibit high test accuracy due to the imbalance in the reference test set of CelebA, which contains many~(84.6\%) samples  of the seen male and female non-blond, compared to that of (14.46\%) challenging unseen female blonds.

\section{Insights}
\mypara{Implicit Classifier Learned by \ourmethod{} is Robust to Subpopulation Shift}
\citeauthor{li2025generative} demonstrate that the implicit classifier, which guides the generation in classifier-free diffusion models remains robust to subpopulation shifts in the absence of spurious labels. However, when spurious labels are available, traditional discriminative baselines that exploit these correlations retain an advantage—evident from the real data setting in \cref{tab:method_comparison}—compared to the performance of implicit diffusion classifier reported in \cref{tab:implicit_method_comparison}.

\ourmethod{} leverages the spurious labels to explicitly enforce conditional independence during training, resulting in a stronger implicit classifier that outperforms diffusion baseline under subpopulation shift~(\cref{tab:implicit_method_comparison}). However, they still trail behind the discriminator based approaches. This improved classifier compared to the standard diffusion models lead to more faithful image generation (\cref{tab:faith-results}), which in turn yields high-utility synthetic data. When used to train downstream classifiers, this data leads to improved robustness under distribution shift (\cref{tab:method_comparison}).

\begin{table}
\centering
\begin{tabular}{lccc}
\toprule
\textbf{Method} & \textbf{Test Acc.} & \textbf{Balanced Acc.} & \textbf{Worst group Acc.} \\
\midrule
 Diffusion & \textbf{61.19}	&	56.47&	33.16 \\
CoInD & 56.39 &			\textbf{60.63 } & \textbf{55.58} \\
\bottomrule
\end{tabular}
\caption{\textbf{Implicit classifier (generative classifier),} $p_{\bm\theta}(\text{blond} = \text{true}\mid \X)$ \textbf{learned by \ourmethod{} achieves better Worst group Acc.} for classifying attribute ``blond'' attribute of CelebA dataset.}
\vspace{-0.5em}
\label{tab:implicit_method_comparison}
\end{table}

\mypara{Qualitatively Accurate Counterfactual Generation}
The ability to compose attributes is reflected in a model’s capacity to generate meaningful counterfactuals—i.e., to answer ``what if?” questions. In our experiments, we generate unseen blond female samples by posing counterfactual queries such as: \textit{What if this non-blond female dyed her hair blond?} or \textit{How would the model interpret a gender change from male to female when generating blond females from blond males?}

As illustrated in \cref{fig:teaser}, Column 1 shows the counterfactual blond female generated from a non-blond female (Column 2). Notably, \ourmethod{} achieves this by modifying only the hair color while preserving all other facial attributes. In contrast, baseline diffusion model tend to generate shorter hair—a spurious correlation from the training distribution where short hair (male feature) correlates with blondes—thus yielding incorrect interpolations. This highlights \ourmethod{}’s superior compositional ability to disentangle and manipulate attributes independently, which translates to a robust downstream classifier.

\begin{figure}[ht]
    \centering
    \vspace{-1.0em}
    \subcaptionbox{\parbox{0.2\textwidth}{ True underlying data distribution}\label{fig:full-2d}}[0.24\textwidth]
    {\includegraphics[width=0.24\textwidth]{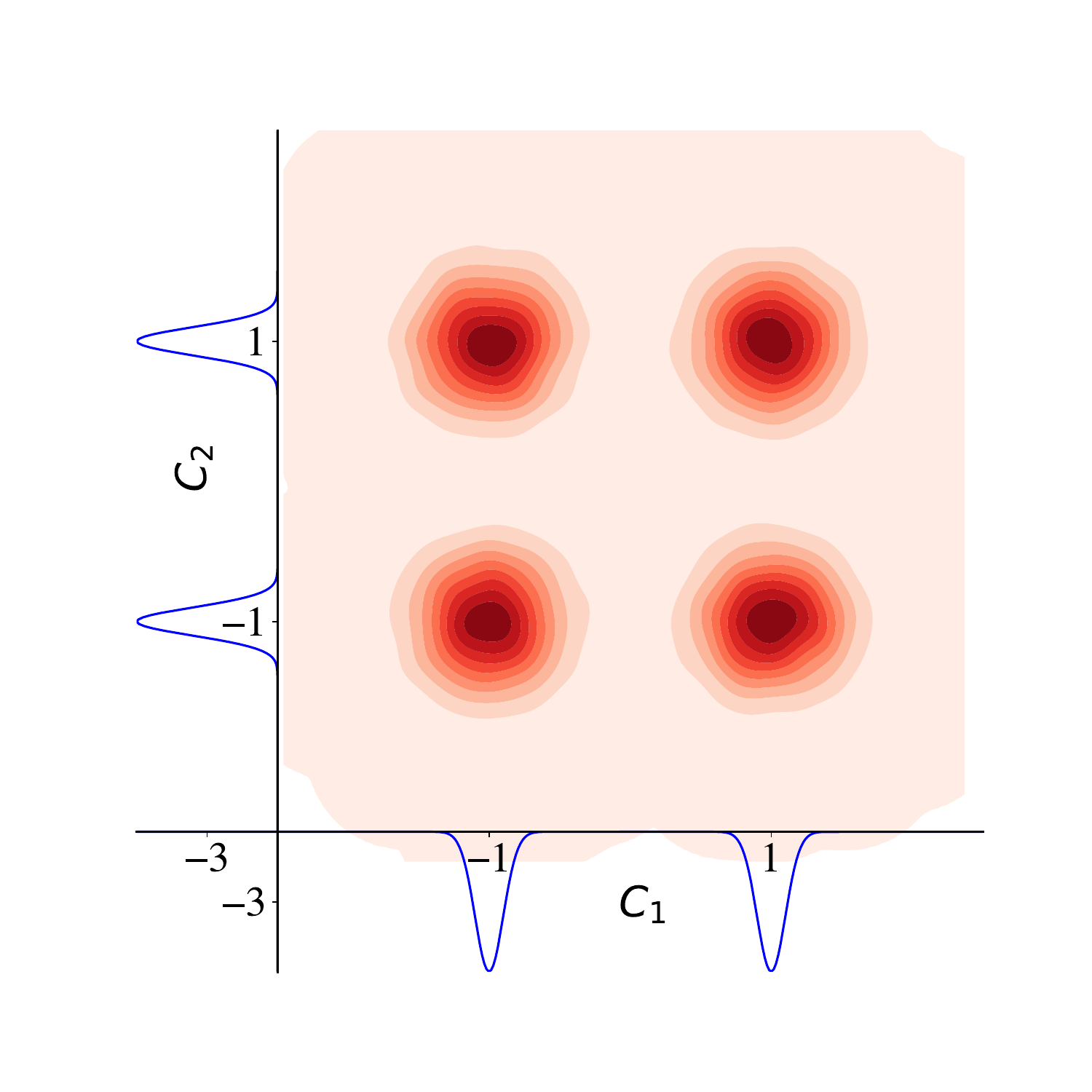}}
    \hfill
    \subcaptionbox{\parbox{0.2\textwidth}{ Training data: Orthogonal~Support}\label{fig:train-2d}}[0.24\textwidth]
    {\includegraphics[width=0.24\textwidth]{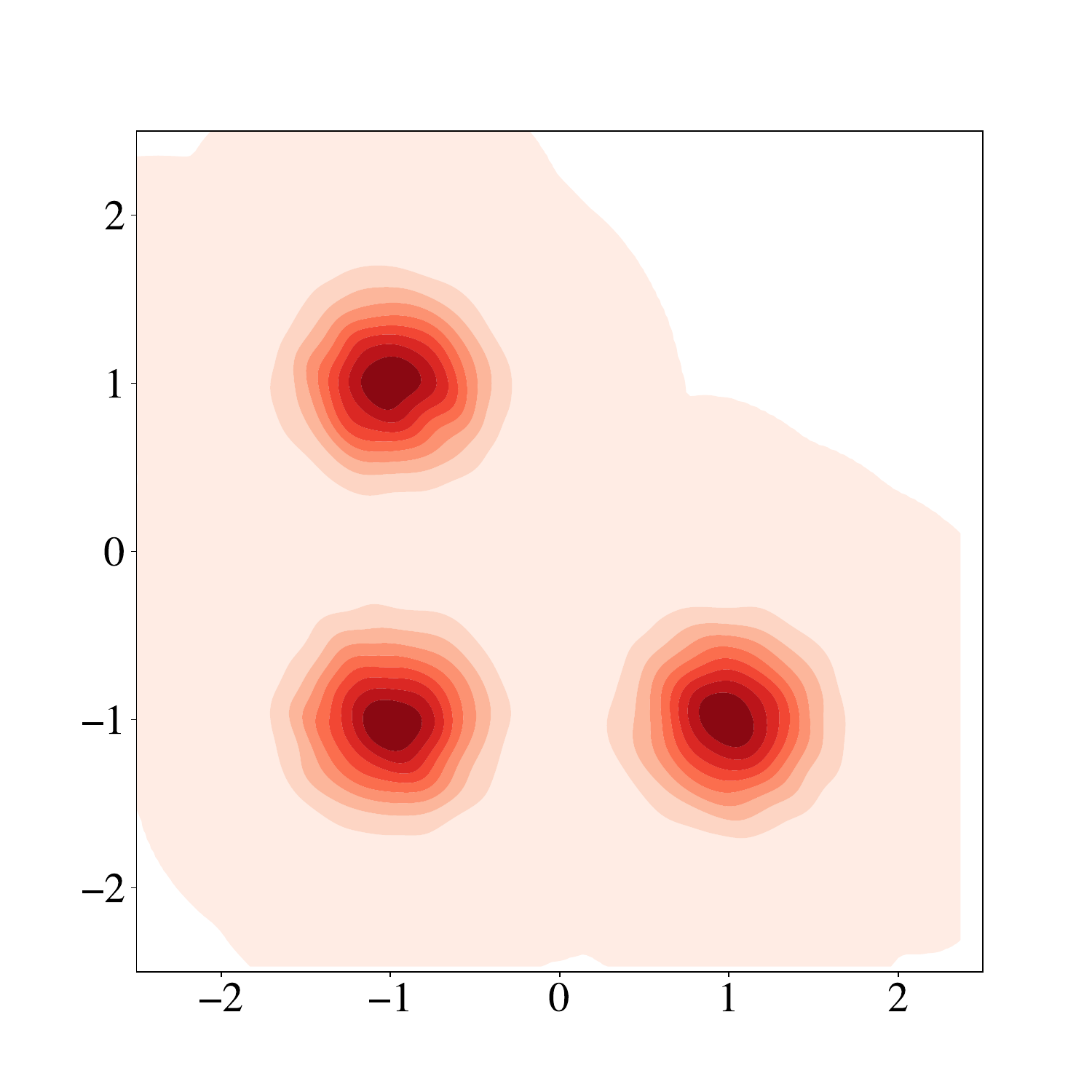}}
    \hfill
    \subcaptionbox{\parbox{0.2\textwidth}{Conditional distribution learned by vanilla diffusion objective\label{fig:vanilla-2d}}}[0.24\textwidth]
    {\includegraphics[width=0.24\textwidth]{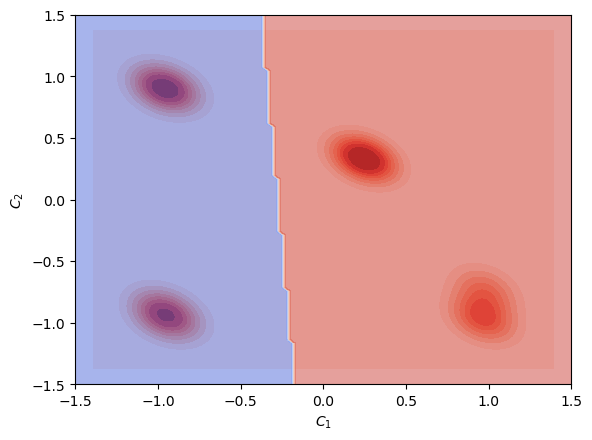}}
    \hfill
    \subcaptionbox{\parbox{0.2\textwidth}{ Conditional distribution learned by \ourmethod{}\label{fig:coind-2d}}}[0.24\textwidth]
    {\includegraphics[width=0.24\textwidth]{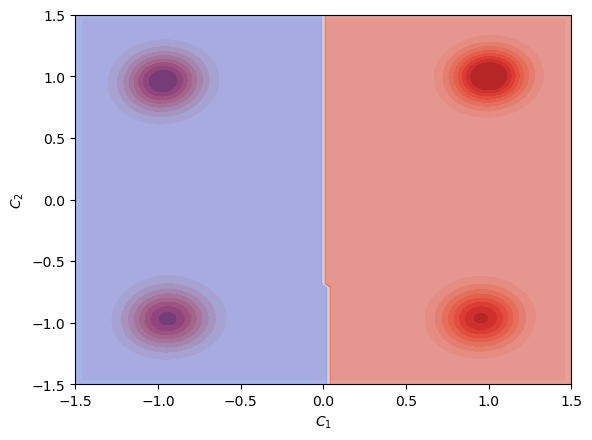}}
    \caption{\ourmethod{} respects underlying independence conditions thereby generating true data distribution (d). }
    \vspace{-1.0em}
    \label{fig:main-2d-gaussian}
\end{figure}

\mypara{ Illustrative Setting: Theoretical Study Using a 2D Gaussian Mixture}  To elucidate the issue of incorrect interpolation, we derive the score for the mixture of Gaussians in closed form. As illustrated in \cref{fig:main-2d-gaussian}, directly learning the marginal 
$p_\theta(X \mid C_0=1)$
from the training distribution (shown in panel (b)) results in a distribution that incorrectly represents a single Gaussian with mean \((1, -1)\) rather than the appropriate mixture comprising Gaussians with means \((1, -1)\) and \((1, 1)\). This error in modeling leads to an inaccurate samples when drawing from $p_\theta(X \mid C_0, C_1=(1,1))$,
as depicted in panel (c). In contrast, \ourmethod{} leverages conditional independence to accurately learn 
$p_\text{train}(X \mid C_0=1)$
ensuring that the resulting distribution is faithful to the true data. Consequently, this approach yields a robust classifier when trained on the correct data representation, as evidenced in panel (d). More details can be found in \cref{appsec:2d-gaussian}.

\section{Related Work}

\mypara{Subpopulation shift} is a long studied topic. Group DRO~\citep{sagawa2019distributionally}, a competitive baseline method, minimizes worst-group error as a proxy for generalization. In contrast, \ourmethod{}, enforces independence between spurious and target features in data generation—a constraint that defines spurious correlation. Compositional Risk Minimization (CRM)~\citep{mahajan2024compositional} is closely related to our work. While CRM enforces compositional constraints on the classifier's output, \ourmethod{} applies these constraints in the data-generating space, where they originate.  Our experimental evaluation directly borrows baseline results from~\citet{mahajan2024compositional}.

\mypara{Synthetic data for improving downstream models} has been discouraged due to the negative evidence from the modal collapse~\cite{alemohammad2023selfconsuminggenerativemodelsmad}. This can be explained because maximum likelihood objective encourages memorization~\cite{kamb2024analytictheorycreativityconvolutional}. Consequently, no new information is provided for classifiers to improve performance. While recent studies \citep{azizi2023synthetic,tian2024stablerep,chen2024human} demonstrated performance improvements using pre-trained generative models for classification. Critically, these approaches often suffer from potential test information leakage, whereas our method achieves robust classifier performance without external information.

\mypara{Compositionality in generative models} Our work concerns compositional generalization in generative models, where the goal is to generate data with unseen attribute compositions. One class of approaches seek to achieve compositionality by combining distinct models trained for each attribute~\citep{du2020compositional, liu2021learning, nie2021controllable, du2023reduce}, which is expensive and scaling linearly with the number of attributes also suffer from incorrect marginals. In contrast, we are interested in monolithic compositional diffusion models that learn compositionality. \citet{liu2023compositional} studied compositionality broadly and proposed methods to represent compositions in terms of marginal probabilities obtained through factorization of the joint distribution. However, these factorized sampling methods fail since the underlying generative model learns inaccurate marginals.

\section{Conclusion}

In this work, we demonstrate that diffusion models trained on partial data, where not all attribute compositions are observed, fail to respect underlying conditional independence relations, thus compromising their ability to exhibit compositionality. To address this issue, we propose a novel synthetic data generation algorithm, \ourmethod{}, which enhances compositional capabilities and allows for enforcing constraints in the data generation process. This approach leads to improved synthetic data quality and enables fine-grained control over the generation process.

The capabilities of \ourmethod{} naturally translate to enhanced robustness of downstream classifiers trained on synthetic data from \ourmethod{} against compositional shift. For a more comprehensive discussion and analysis of \ourmethod{}, including 2D Gaussian experiments~\cref{appsec:2d-gaussian}, extension to flow-based models~\cref{appsec:flow}, and its limitations~\cref{appsec:limitations}, readers are directed to the appendix.

\bibliography{iclr2025_conference}
\bibliographystyle{iclr2025_conference}
\newpage
\appendix
\addcontentsline{toc}{section}{Appendix} 
\part{Appendix} 
\parttoc 

\newpage

\section{Preliminaries of Score-based Models\label{appsec:preliminaries}}

\paragraph{Score-based models}
Score-based models~\citep{song2021scorebased} learn the score of the observed data distribution, $P_\text{train}(\X)$ through score matching~\citep{hyvarinen2005estimation}. The score function $s_{\bm{\theta}}(\mathbf{x}) = \nabla_{\mathbf{x}} \log p_{\bm{\theta}}(\mathbf{x})$ is learned by a neural network parameterized by $\bm{\theta}$.
\begin{equation}
    L_{\text{score}} = \mathbb{E}_{\mathbf{x} \sim p_\text{train}} \left[ \left\| s_{\bm{\theta}}(\mathbf{x}) - \nabla_{\mathbf{x}} \log p_{\text{train}}(\mathbf{x}) \right\|_2^2 \right]
\label{score-matching}\end{equation}
During inference, sampling is performed using Langevin dynamics:
\begin{equation} \label{og_sampling}
    \mathbf{x}_t = \mathbf{x}_{t-1} + \frac{\eta}{2} \nabla_{\mathbf{x}} \log p_{\bm{\theta}}(\mathbf{x}_{t-1}) + \sqrt{\eta} \epsilon_t, \quad \epsilon_t \sim \mathcal{N}(0,1)
\end{equation}
where \( \eta > 0 \) is the step size. As \( \eta \rightarrow 0 \) and \( T \rightarrow \infty \), the samples \( \mathbf{x}_t \) converge to \( p_{\bm{\theta}}(\X) \) under certain regularity conditions~\citep{welling2011bayesian}.

\mypara{Diffusion models}\cite{song2019generative} proposed a scalable variant that involves adding noise to the data \cite{ho2020denoising} has shown its equivalence to Diffusion models. Diffusion models are trained by adding noise to the image $\mathbf{x}$ according to a noise schedule, and then neural network, $\epsilon_\theta$ is used to predict the noise from the noisy image, $\mathbf{x_t}$. The training objective of the diffusion models is given by:
\begin{equation}\label{eq:og_score}
L_{\text{score}} = \E_{\mathbf{x}\sim p_\text{train}}\E_{t \sim [0,T]}\left\| \mathbf{\epsilon} - \epsilon_\theta \left(\mathbf{x}_t, t \right) \right\|^2
\end{equation}
Here, the perturbed data \( \mathbf{x}_t \) is expressed as: $\mathbf{x}_t = \sqrt{\bar{\alpha}_t}\mathbf{x} + \sqrt{1-\bar{\alpha}_t}\mathbf{\epsilon}$
where \(\bar{\alpha}_t = \prod_{i=1}^T \alpha_i\), for a pre-specified noise schedule \(\alpha_t\).
The score can be obtained using, 
\begin{equation}
    s_\theta(\mathbf{x}_t, t) \approx  -\frac{\epsilon_\theta(\mathbf{x}_t, t)}{\sqrt{1 - \bar{\alpha}_t}}\label{diffusion-as-score}
\end{equation}
Langevin dynamics can be used to sample from the $s_\theta(\mathbf{x}_t, t)$ to generate samples from $p(\X)$. The conditional score~\citep{dhariwal2021diffusion} is used to obtain samples from the conditional distribution \( p_{\bm{\theta}}(\X \mid C) \) as:
\begin{align*}
   \nabla_{\X_t} \log p(\X_t \mid C) &=  \underbrace{\nabla_{\X_t} \log p_{\bm{\theta}}(\X_t)}_{\text{Unconditional score}} + \gamma \nabla_{\X_t} \log \underbrace{p_{\bm{\theta}}(C \mid \X_t)}_{\text{noisy classifier}}
\end{align*}
where $\gamma$ is the classifier strength. Instead of training a separate noisy classifier, \citeauthor{ho2022classifier} have extended to conditional generation by training $ \nabla_{\X_t} \log p_{\bm{\theta}}(\X_t \mid C)= s_\theta(\X_t, t,C)$. The sampling can be performed using the following equation:
\begin{equation}\label{classifier_free_sampling}
\nabla_{\X_t} \log p(\X_t\mid C) = (1-\gamma)\nabla_{\X_t} \log p_{\bm{\theta}}(\X_t) + \gamma \nabla_{\X_t} \log p_{\bm{\theta}}(\X_t\mid C)
\end{equation}
However, the sampling needs access to unconditional scores as well. Instead of modelling $\nabla_{\X_t} \log p_{\bm{\theta}}(\X_t)$,  $\nabla_{\X_t} \log p_{\bm{\theta}}(\X_t | C)$ as two different models \citeauthor{ho2022classifier} have amortize training a separate classifier training a conditional model $s_\theta(\mathbf{x}_t, t,\mathbf{c})$ jointly with unconditional model trained by setting $c=\varnothing$. 

In the general case of classifier-free guidance, a single model can be effectively trained to accommodate all subsets of attribute distributions. During the training phase, each attribute $c_i$ is randomly set to $\varnothing$ with a probability $p_{\text{uncond}}$. This approach ensures that the model learns to match all possible subsets of attribute distributions. Essentially, through this formulation, we use the same network to model all the possible subsets of conditional probability. 

Once trained, the model can generate samples conditioned on specific attributes, such as $c_i$ and $c_j$, by setting all other conditions to $\varnothing$. The conditional score is then computed as, $\nabla_{\X_t} \log p_{\bm{\theta}}(\X_t | c_i, c_j) = s_\theta(\mathbf{x}_t, \mathbf{c}^{i,j})$, where $\mathbf{c}^{i,j}$ represents the condition vector with all values other than $i$ and $j$ set to $\varnothing$. This method allows for flexible and efficient sampling across various attribute combinations.

\mypara{Estimating Guidance}
Once the diffusion model is trained, we investigate the implicit classifier, $p_\theta(C|\X)$, learned by the model. This will give us insights into the learning process of the diffusion models. \citep{Li_2023_ICCV} have shown a way to calculate $p_\theta(C_i=c_i\mid \X=\mathbf{x})$, borrowing equation (5), (6) from \citep{Li_2023_ICCV}.
\begin{align*}
    p_\theta(C_i=c_i \mid \mathbf{x}) &= \frac{p(c_i) \, p_\theta(\mathbf{x} \mid c_i)}{\sum_k p(c_k) \, p_\theta(\mathbf{x} \mid c_k)} 
\end{align*}
\begin{equation}\label{eq:marginal}
    p_\theta(C_i=c_i \mid \mathbf{x}) = \frac{\exp\{-\mathbb{E}_{t,\epsilon}[\| \epsilon - \epsilon_\theta(\mathbf{x}_t,t, \mathbf{c}^i) \|^2]\}}{\E_{C_i} \left[ \exp\{-\mathbb{E}_{t,\epsilon}[\| \epsilon - \epsilon_\theta(\mathbf{x}_t,t, \mathbf{c}^i) \|^2]\}\right]}
\end{equation}

Likewise, we can extend it to joint distribution by 
\begin{align}\label{eq:diff-model-classifier}
    p_\theta(C_i=c_i,C_j=c_j \mid  \mathbf{x}) &= \frac{\exp\{-\mathbb{E}_{t,\epsilon}[\| \epsilon - \epsilon_\theta(\mathbf{x}_t,t, \mathbf{c}^{i,j}) \|^2]\}}{\E_{C_i,C_j} \left[ \exp\{-\mathbb{E}_{t,\epsilon}[\| \epsilon - \epsilon_\theta(\mathbf{x}_t,t,\mathbf{c}^{i,j}) \|^2]\}\right] }
\end{align}
\mypara{Practical Implementation} The authors~\citeauthor{Li_2023_ICCV}. have showed many axproximations to compute $\mathbb{E}_{t,\epsilon}$. However, we use a different approximation inspired by \cite{kynkaanniemi2024applying}, where we sample 5 time-steps between [300,600] instead of these time-steps spread over the [0, T].

\subsection{Computing JSD\label{sec:jsd}}

We are interested in understanding the causal structure learned by diffusion models. Specifically, we aim to determine whether the learned model captures the conditional independence between attributes, allowing them to vary independently. This raises the question: \emph{Do diffusion models learn the conditional independence between attributes?} The conditional independence is defined by:
\begin{equation}\label{eq:ci}
p_{\bm{\theta}}(C_i,C_j \mid \X) = p_{\bm{\theta}}(C_i \mid \X)p_{\bm{\theta}}(C_j \mid \X)
\end{equation}
We aim to measure the violation of this equality using the Jensen-Shannon divergence (JSD) to quantify the divergence between two probability distributions:
\begin{equation}\label{eq:newwasser}
\jsd = \mathbb{E}_{p_{\text{data}}}\left[ D_{\text{JS}}\left( p_{\bm{\theta}}(C \mid \X) \mid\mid p_{\bm{\theta}}(C_i \mid \X)p_{\bm{\theta}}(C_j \mid \X) \right) \right]
\end{equation}
The joint distribution, $p_{\bm{\theta}}(C_i, C_j \mid \X)$, and the marginal distributions, $p_{\bm{\theta}}(C_i \mid \X)$ and $p_{\bm{\theta}}(C_j \mid \X)$, are evaluated at all possible values that $C_i$ and $C_j$ can take to obtain the probability mass function (pmf). The probability for each value is calculated using Equation~\cref{eq:diff-model-classifier} for the joint distribution and Equation~\cref{eq:marginal} for the marginals.

\paragraph{Practical Implementation} For the diffusion model with multiple attributes, the violation in conditional \textit{mutual} independence should be calculated using all subset distributions. However, we focus on pairwise independence. We further approximate this in our experiments by computing $\jsd$ between the first two attributes, $C_1$ and $C_2$. We have observed that computing $\jsd$ between any attribute pair does not change our examples' conclusion.

\subsection{Conformity Score (CS) \label{cs-score}}

To measure the CS, we first infer attributes $(\hat{c}_1, \dots, \hat{c}_n)$ from the generated images $\hat{\X}$ using attribute-specific classifiers $\phi_{C_i}$ and compare them against the expected attributes from the input  composition $(c_1,\dots, c_n)$. We refer to this accuracy as \emph{conformity score}~(CS) and is given by 
\begin{equation}
    \cs(g) = \mathbb{E}_{p(C)p(U)}\left[\prod_i^n \mathds{1}(C_i, \phi_{C_i}(g(C, U))) \right]
\end{equation}
where $\mathds{1}(\cdot, \cdot)$, $g$, and $U$ are the indicator function, diffusion model, and the stochastic noise in the generation process respectively

To obtain a attribute-specific classifier, we train a single ResNet-18~\citep{he2016deep} classifier with multiple classification heads, one corresponding to each attribute, and trained on the full support. The effectiveness of the classifier in predicting the attributes is reported in~\cref{sec:cs-acc}

\section{Proofs for Claims}

In this section, we detail the mathematical derivations for violation of conditional independence in diffusion models in \cref{appsubsec:incorrect-marginals}, and then derive the final loss function of \ourmethod{} in \cref{sec:proof}.

\subsection{Standard diffusion model objective is not suitable for compositionality\label{appsubsec:incorrect-marginals}}

This section proves that the violation in conditional independence in diffusion models is due to learning incorrect marginals, $p_\text{train}(\X \mid C_i)$ under $C_i \notind C_j$.
We leverage the causal invariance property: $p_{\text{train}}(\X \mid C) = p_{\text{true}}(\X \mid C) $, where $p_{\text{train}}$ is the training distribution and $p_{\text{true}}$ is the true underlying distribution.


Consider the training objective of the score-based models in classifier free formulation~\cref{score-matching}. For the classifier-free guidance, a single model $s_{\bm{\theta}}(\mathbf{x}, C)$ is effectively trained to match the score of all subsets of attribute distributions. Therefore, the effective formulation for classifier-free guidance can be written as,
\begin{equation}
    L_{\text{score}} = \mathbb{E}_{\mathbf{x} \sim p_\text{train}} \mathbb{E}_{S} \left[ \left\| \nabla_{\mathbf{x}} \log p_{\bm{\theta}}(\mathbf{x} \mid c_S) - \nabla_{\mathbf{x}} \log p_{\text{train}}(\mathbf{x}\mid c_S) \right\|_2^2 \right]\label{eq:fishdiv}
\end{equation}
where $S$ is the power set of attributes.

From the properties of Fisher divergence, $L_{\text{score}}=0$ iff $p_{\bm{\theta}}(\X \mid c_S) = p_\text{train}(\X \mid c_S)$, $\forall S$. In the case of marginals, $p_{\bm{\theta}}(\X \mid C_i)$ i.e. $S = \{C_i\}$ for some $1\leq i \leq n$,
\begin{align*}
   p_{\bm{\theta}}(\X \mid C_i) &=  p_{\text{train}}(\X\mid C_i) \\
   &= \sum_{C_{-i}} p_{\text{train}}(\X\mid C_i, C_{-i})p_{\text{train}}(C_{-i}\mid C_i) \\ 
   &= \sum_{C_{-i}} p_{\text{true}}(\X\mid C_i, C_{-i})p_{\text{train}}(C_{-i}\mid C_i) \\ 
   &\neq \sum_{C_{-i}} p_{\text{true}}(\X\mid C_i, C_{-i})p_\text{true}(C_{-i}) = p_{\text{true}}(\X\mid C_i) \\
   \implies p_{\bm{\theta}}(\X \mid C_i) &\neq p_{\text{true}}(\X\mid C_i) \numberthis\label{appeq:marginal-ineq}
\end{align*}
Where $C_{-i} = \prod_{\substack{j=1 \\ j \neq i}}^{n} C_j$, which is every attribute except $C_i$.
Therefore, the objective of the score-based models is to maximize the likelihood of the marginals of training data and not the true marginal distribution, which is different from the training distribution when $C_i \notind C_j$.

\subsection{Step-by-step derivation of \ourmethod{} in \cref{sec:method}\label{sec:proof}}

The objective is to train the model by explicitly modeling the joint likelihood following the causal factorization. The minimization for this objective can be written as,
\begin{align}
    \mathcal{L}_{\text{comp}} = \mathcal{W}_2\left( p(\X \mid C), \frac{p_{\bm\theta}(\X)}{p_{\bm\theta}(C)} \prod_i \frac{p_{\bm\theta}(\X \mid C_i)p_{\bm\theta}(C_i)}{p_{\bm\theta}(\X)} \right) \label{eq:linv-original-2}
\end{align}
where $\mathcal{W}_2$ is 2-Wasserstein distance. Applying the triangle inequality to \cref{eq:linv-original-2} we have,
\begin{align}
    \mathcal{L}_{\text{comp}} \leq \underbrace{\mathcal{W}_2\left( p(\X \mid C), p_{\bm\theta}(\X \mid C) \right)}_{\text{Distribution matching}} + \underbrace{\mathcal{W}_2\left(p_{\bm\theta}(\X \mid C), \frac{p_{\bm\theta}(\X)}{{p_{\bm\theta}(C)}} \prod_i^n \frac{p_{\bm\theta}(\X \mid C_i)p_{\bm\theta}(C_i)}{p_{\bm\theta}(\X)} \right)}_{\text{\quad Conditional Independence}} \label{eq:2-linv-triangle}
\end{align}
\citep{kwon2022scorebased} showed that under some conditions, the Wasserstein distance between $p_0(\X), q_0(\X)$ is upper bounded by the square root of the score-matching objective. Rewriting Equation 16 from \citep{kwon2022scorebased}
\begin{align}
    \mathcal{W}_2\left( p_0(\X), q_0(\X) \right) \leq K\sqrt{\E_{p_0({\X})}\left[||\nabla_{\X} \log p_0(\X) - \nabla_{\X} \log q_0(\X)||_2^2\right] } \label{kwon}
\end{align}

\mypara{Distribution matching}Following \cref{kwon} result, the first term in \cref{eq:2-linv-triangle}, replacing $p_0$ as $p$ and  $q_0$ as $p_\theta$ will result in
\begin{align*}
    \mathcal{W}_2\left( p(\X \mid C), p_{\bm\theta}(\X \mid C)\right) &\leq K_1 \sqrt{\E_{p_0({\X})}\left[||\nabla_{\X} \log p(\X \mid C) - \nabla_{\X} \log p_\theta(\X)||_2^2\right] } \\
    &= K_1 \sqrt{\mathcal{L}_{\text{score}} } \numberthis \label{eq:sm}
\end{align*}
\mypara{Conditional Independence}Following \cref{kwon} result, the second term in \cref{eq:2-linv-triangle}, replacing $p_0$ as $p_\theta$ and $q_0(\X)$ as $\frac{p_{\bm\theta}(\X)}{{p_{\bm\theta}(C)}} \prod_i^n \frac{p_{\bm\theta}(\X \mid C_i)p_{\bm\theta}(C_i)}{p_{\bm\theta}(\X)}$
\begin{align*}
\mathcal{W}_2\left(p_{\bm\theta}(\X \mid C), \frac{p_{\bm\theta}(\X)}{{p_{\bm\theta}(C)}} \prod_i^n \frac{p_{\bm\theta}(\X \mid C_i)p_{\bm\theta}(C_i)}{p_{\bm\theta}(\X)} \right) \\
    \le \sqrt{\E\| \score p_\theta(\X\mid C) - \score \frac{p_{\bm\theta}(\X)}{{p_{\bm\theta}(C)}} \prod_i^n \frac{p_{\bm\theta}(\X \mid C_i)p_{\bm\theta}(C_i)}{p_{\bm\theta}(\X)} \|_2^2}
\end{align*}
Further simplifying and incorporating $\score p_{\theta}(C_i)=0$ and $\score p_{\theta}(C)=0$ will result in 
\begin{align*}
& \mathcal{W}_2\left(p_{\bm\theta}(\X \mid C), \frac{p_{\bm\theta}(\X)}{{p_{\bm\theta}(C)}} \prod_i^n \frac{p_{\bm\theta}(\X \mid C_i)p_{\bm\theta}(C_i)}{p_{\bm\theta}(\X)} \right) \\
    &\le K_2 \sqrt{\underbrace{\E\| \score p_\theta(\X\mid C) - \score p_\theta(\X) - \sum_i \left[\score p_\theta(\X\mid C_i) - \score p_\theta(\X)\right]\|_2^2}_{\lci}}  \\
    &= K_2 \sqrt{\lci}
    \numberthis \label{eq:cond}
\end{align*}

Substituting \cref{eq:sm}, \cref{eq:cond} in \cref{eq:2-linv-triangle} will result in our final learning objective
\begin{equation}
    \mathcal{L}_{\text{comp}} \leq K_1\sqrt{\mathcal{L}_{\text{score}}} + K_2\sqrt{\mathcal{L}_{\text{CI}}}
    \label{eq:lcomp_2}
\end{equation}
where $K_1,K_2$ are positive constants, i.e., the conditional independence objective $\lci$ is incorporated alongside the existing score-matching loss $\mathcal{L}_{\text{score}}$.

Note that \cref{eq:cond} is the Fisher divergence between the joint $p_\theta(\X\mid C)$ and the causal factorization $\frac{p_\theta(X)}{p_\theta(C)}\prod_i \frac{p_\theta(\X \mid C_i)p_\theta(C_i)}{p_\theta(\X)}$. From the properties of Fisher divergence~\citep{sanchez2012jensen}, $\lci = 0$ iff $p_{\mathbb{\theta}}(\X \mid C) = \frac{p_{\mathbb{\theta}}(\X)}{p_{\mathbb{\theta}}(C)}\prod_i^n\frac{p_{\mathbb{\theta}}(\X\mid C_i)p_{\mathbb{\theta}}(C_i)}{p_{\mathbb{\theta}}(\X)}$ and further implying, $\prod_i p_\theta(C_i\mid\X) = p_\text{train}(C\mid \X)$

When $L_{\text{comp}} = 0$: $P_\theta(\X\mid C) = P_\text{train}(\X \mid C) = P(\X \mid C)$, and $\prod_i p_\theta(C_i\mid\X) = p_\text{train}(C\mid \X)$. This implies that the learned marginals obey the causal independence relations from the data-generation process, leading to more accurate marginals.

\subsection{Connection to compositional generation from first principles }\label{sec:connect}

Compositional generation from first principles~\cite{wiedemer2024compositional} have shown that restricting the function to a certain compositional form will perform better than a single large model. In this section, we show that, by enforcing conditional independence, we restrict the function to encourage compositionality.

Let \( c_1, c_2, \ldots, c_n\) be independent components such that \( c_1, c_2, \ldots, c_n\in \mathbb{R} \). Consider an injective function \( f: \mathbb{R}^n \to \mathbb{R}^d \) defined by \( f(c) = x \). If the components, $c$ are conditionally independent given \( x \) the cumulative functions, $F$ must satisfy the following constraint:
\begin{equation}
F_{C_i,C_j,\dots, C_n\mid X=x}(c_i,c_j,\ldots,c_n) = \prod_i F_{C_i\mid X=x}(c_i)
\end{equation}
$F^{-1}_{C_i,C_j,\dots, C_n\mid X=x}(x) = \inf\{c_i,c_j,\ldots,c_n \mid  F(c_i,c_j,\ldots,c_n)\ge x\}$, where $F^{-1}_{c_i,c_j,\dots, C_n\mid X=x}$ is a generalized inverse distribution function.
\begin{align*}
f(c_i,c_j,\ldots,c_n) &= (f \circ F^{-1}_{c_i,c_j,\dots, C_n\mid X=x})(\prod_i F_{C_i\mid X=x}(c_i)) \\
&= (f \circ F^{-1}_{c_i,c_j,\dots, C_n\mid X=x}\circ 
  e)(\sum_i 
  \log F_{C_i\mid X=x}(c_i)) \\
&=g(\sum_i \phi_i(c_i)) \\
\end{align*}
Therefore, we are restricting $f$ to take a certain functional form. However, it is difficult to show that the data generating process, $f$, meets the rank condition on the Jacobian for the sufficient support assumption~\cite{wiedemer2024compositional}, which is also the limitation discussed in their approach. Therefore, we cannot provide guarantees. However, this section provides a functional perspective of \ourmethod{}.

\section{Practical Considerations\label{sec:practical}}

To facilitate scalability and numerical stability for optimization, we introduce two approximations to the upper bound of our objective function~\cref{eq:lcomp}.

\subsection{Scalability of $\lci$\label{appsubsec:scalable-lci}}

A key computational challenge posed by~\cref{eq:cond-ind-score} is that the number of model evaluations grows linearly with the number of attributes. The \cref {eq:cond-ind-score} is derived from conditional independence formulation as follows:
\begin{equation}
    p_\theta(C \mid X) = \prod_i p_\theta(C_i \mid X)\label{appeq:ci}.
\end{equation}
By applying Bayes' theorem to all terms, we obtain,
\begin{equation}
   \frac{p_\theta(\X \mid C)p_\theta(C)}{p_\theta(X)}  = \prod_i \frac{p_\theta(\X \mid C_i)p_\theta(C_i)}{p_\theta(\X)}
\end{equation}
Note that this formulation is equal to the causal factorization. From this, by applying logarithm and differentiating w.r.t. $\X$, we derive the score formulation.
\begin{align}
    \nabla_{\X} \log p_\theta(\X\mid C) = \nabla_{\X}\log  \sum_i p_\theta(\X \mid C_i) - \nabla_{\X} \log  p_\theta(\X)  \label{eq:cdform}
\end{align}
The $L_2$ norm of the difference between LHS and RHS of the objective in~\cref{eq:cdform} is given by, which forms our $\lci$ objective.
\begin{align}
    \lci = \lVert\nabla_{\X} \log p_\theta(\X\mid C) - \left(\nabla_{\X}\log  \sum_i p_\theta(\X \mid C_i) - \nabla_{\X} \log  p_\theta(\X)\right)\rVert_2^2 \label{eq:lci}
\end{align}
Due to the $\sum_i$, in the equation, the number of model evaluations grows linearly with the number of attributes~($n$). This $\mathcal{O}(n)$ computational complexity hinders the approach's applicability at scale. To address this, we leverage the results of~\citep{hammond2006essential}, which shows conditional independence is equivalent to pairwise independence under large $n$ to reduce the complexity to $\mathcal{O}(1)$ in expectation. This allows for a significant improvement in scalability while maintaining computational efficiency. Using this result, we modify~\cref{appeq:ci} to: 
\[
    p_\theta(C_i, C_j \mid X) = p_\theta(C_i \mid X)p_\theta(C_j \mid X). \quad \forall i,j
\]
Accordingly, we can simplify the loss function for conditional independence as follows:
\begin{align}
    \lci &= \E_{p({\X,C})}\E_{j,k} \lVert\nabla_{X} [ \log p_\theta(X|C_j,C_k) -  \log p_\theta(X|C_j) -  \log p_\theta(X|C_k) +  \log p_\theta(X)]  \rVert_2^2
    \label {eq:lci_3}.
\end{align}
In score-based models, which are typically neural networks, the final objective is given as:
\begin{equation}\label{eq:reduced_eq}
    \mathcal{L}_{\text{CI}} = \E_{p({\X,C})}\E_{j,k}\lVert s_{\bm\theta}(\X, C_j,C_k)  - s_{\bm\theta}(\X, C_j) - s_{\bm\theta}(\X, C_k) + s_{\bm\theta}(\X, \varnothing) \rVert_2^2
\end{equation}
where $s_{\bm\theta}(\cdot) \coloneqq\nabla_{\X} \log p_\theta(\cdot)$ is the score of the distribution modeled by the neural network. We leverage classifier-free guidance to train the conditional score $s_{\bm\theta}(\X, C_i)$ by setting $C_k = \varnothing$ for all $k \neq i$, and likewise for $s_{\bm\theta}(\X, C_i, C_j)$, we set $C_k=\varnothing$ for all $k \not \in \{i,j\}$.

\subsection{Simplification of Theoretical Loss\label{appsubsec:theory-loss-simple}}

In \cref{eq:lcomp}, we showed that the 2-Wassertein distance between the true joint distribution $p(\X\mid C)$ and the causal factorization in terms of the marginals $p(\X\mid C_i)$ is upper bounded by the weighted sum of the square roots of $\lscore$ and $\lci$ as $\mathcal{L}_{\text{comp}} \leq K_1\sqrt{\mathcal{L}_{\text{score}}} + K_2\sqrt{\mathcal{L}_{\text{CI}}}$. In practice, however, we minimized a simple weighted sum of $\lscore$ and $\lci$, given by $\mathcal{L}_{\text{final}} = \lscore + \lambda\lci$ as shown in \cref{eq:final_eq} instead of \cref{eq:lcomp}. We used \cref{eq:final_eq} to avoid the instability caused by larger gradient magnitudes (due to the square root). \cref{eq:final_eq} also provided the following practical advantages: \textbf{(1)}~the simplicity of the loss function that made hyperparameter tuning easier, and \textbf{(2)}~the similarity of \cref{eq:final_eq} to the loss functions of pre-trained diffusion models allowing us to reuse existing hyperparameter settings from these models. 

\subsection{Choice of Hyperparameter $\lambda$ \label{sec:hyperparameter}}

The value for the hyperparameter $\lambda$ is chosen such that the gradients from the score-matching objective $L_{\text{score}}$ and the conditional independence objective $L_{\text{CI}}$ are balanced in magnitude. One way to choose $\lambda$ is by training a vanilla diffusion model and setting $\lambda$ = $\frac{L_{score}}{L_{CI}}$. As a rule of thumb, we recommend the simplified setting: $\lambda = L_{score}\times 4000$.

\subsection{Limitations\label{appsec:limitations}}

This paper considered compositions of a closed set of attributes. As such, \ourmethod{} requires pre-defined attributes and access to data labeled with the corresponding attributes. Moreover, \ourmethod{} must be enforced during training, which requires retraining the model whenever the attribute space changes to include additional values. Instead, state-of-the-art generative models seek to operate without pre-defined attributes or labeled data and generate open-set compositions. Despite the seemingly restricted setting of our work, our findings provide valuable insights into a critical limitation of current generative models, namely their failure to generalize for unseen compositions, by identifying the source of this limitation and proposing an effective solution to mitigate it.

\section{Experiment Details\label{appsec:experiment-details}}

\subsection{\ourmethod{} Algorithm\label{sec:algo}}

\algrenewcommand\algorithmicindent{0.5em}
\begin{algorithm}[ht]
    \caption{\ourmethod{} Training} \label{alg:training}
    \small
    \begin{algorithmic}[1]
      \Repeat
        \State $(\mathbf{c}, \mathbf{x}_0 ) \sim p_{\text{train}}(\mathbf{c},x)$
        \State \hlight{$c_k \leftarrow \varnothing \text{ with probability } p_{uncond}$} \Comment{ Set element of index,$k$ i.e, $c_k$ to $\varnothing$ with $p_{uncond} \forall k \in [0,N]$ probability}
        \State \hlight{$i \sim  \mathrm{Uniform}(\{0, \ldots, N\}), j \sim  \mathrm{Uniform}(\{0, \ldots, N\} \setminus \{i\})$}  \Comment{Select two random attribute indices}
        \State $t \sim \mathrm{Uniform}(\{1, \dotsc, T\})$
        \State $\boldsymbol{\epsilon}\sim\mathcal{N}(\mathbf{0},\mathbf{I})$
        \State $x_t = \sqrt{\bar\alpha_t} \mathbf{x}_0 + \sqrt{1-\bar\alpha_t}\boldsymbol{\epsilon}$
        \State \hlight{$ \mathbf{c}^i, \mathbf{c}^j, \mathbf{c}^{i,j} \leftarrow \mathbf{c}$}
        \State \hlight{$\mathbf{c}^i \leftarrow \{c_k = \varnothing \mid k \ne i\}$, $\mathbf{c}^j \leftarrow \{c_k = \varnothing \mid k \ne j\}$, $\mathbf{c}^{i,j} \leftarrow \{c_k = \varnothing \mid k \not \in  \{i,j\}\}$, $\mathbf{c}^{\varnothing} \leftarrow \varnothing$} 
        \State \hlight{$L_{CI} =  ||   \boldsymbol{\epsilon}_\theta(\mathbf{x}_t, t,\mathbf{c}^i) + \boldsymbol{\epsilon}_\theta(\mathbf{x}_t, t,\mathbf{c}^j) - \boldsymbol{\epsilon}_\theta(\mathbf{x}_t, t,\mathbf{c}^{i,j}) -   \boldsymbol{\epsilon}_\theta(\mathbf{x}_t, t,\mathbf{c}^{\varnothing}) ||_2^2$}
        \State Take gradient descent step one
        \Statex $\qquad \nabla_\theta [ \left\| \boldsymbol{\epsilon} - \boldsymbol{\epsilon}_\theta(\mathbf{x}_t, t,\mathbf{c}) \right\|^2$ \hlight{$+\lambda L_{CI}$} $]$
      \Until{converged}
    \end{algorithmic}
\end{algorithm}
To compute pairwise independence in a scalable fashion, we randomly select two attributes, $i$ and $j$, for a sample in the batch and enforce independence between them. As the score in \cref{diffusion-as-score} is given by $\frac{\epsilon_\theta(\mathbf{x}_t, t)}{\sqrt{1 - \bar{\alpha}_t}}$. The final equation for enforcing $\lci$ will be:
\begin{equation*}
L_{CI} = \frac{1}{1 - \bar{\alpha}_t} \left\| \boldsymbol{\epsilon}_\theta(\mathbf{x}_t, t,\mathbf{c}^i) + \boldsymbol{\epsilon}_\theta(\mathbf{x}_t, t,\mathbf{c}^j) - \boldsymbol{\epsilon}_\theta(\mathbf{x}_t, t,\mathbf{c}^{i,j}) - \boldsymbol{\epsilon}_\theta(\mathbf{x}_t, t,\mathbf{c}^{\varnothing}) \right\|_2^2
\end{equation*}
We follow \cite{ho2020denoising} to weight the term by $1 - \bar{\alpha}_t$. This results in an algorithm for \ourmethod{}, requiring only a few modifications of lines from \citep{ho2022classifier}, highlighted below.
\mypara{Practical Implementation} In our experiments, we have used $p_\text{uncond} = 0.3$

\subsection{Training details, Architecture, and Sampling\label{training-details}}

To generate CelebA images, we scale the image size to $64\times 64$. We use the latent encoder of Stable Diffusion 3 (SD3)~\citep{esser2024scaling} to encode the images to a latent space and perform diffusion in the latent space. The model uses an AdamW optimizer with a learning rate of $1.0\times10^{-4}$ and trains for 500,000 steps. It employs a DDPM train noise scheduler with a cosine noise schedule and 1000 train noise steps. The architecture is based on a U-Net model with 2 layers per block and, with block out channels of [224, 448, 672, 896]. It has an attention head dimension of 8, and 8 norm groups. trained on a A6000 GPU.

\mypara{Sampling}

To generate samples for a composition, we sample from the referenced equation~\cref{eq:and-score-2} using $\gamma=0.46$ to achieve more pronounced gender features, employing the Denoising Diffusion Implicit Models (DDIM)~\citep{song2020denoising} for 250 steps. We apply identical sampling settings across both standard diffusion models and our proposed method. This configuration was selected because it consistently produces high-fidelity samples, demonstrating robust performance across both model architectures. By maintaining consistent sampling parameters, we ensure a fair and comparable evaluation of the generative capabilities of \ourmethod{}.

\mypara{Evaluation}: We evaluate our method by computing the Fr\'echet Inception Distance (FID). Specifically, we sample 5,000 generated images of unseen blond females following method described above and compare them against a reference set of celebA blond female images.

\subsection{Accuracy of classifiers for Conformity Score (CS) \label{sec:cs-acc}}

\begin{table}[h]
\caption{ResNet-18 accuracy of classifying attributes on CelebA Dataset}
\label{tab:summary-cmnist}
\begin{tabular}{llcc}
    \toprule
    Feature & Attributes & Possible Values & Accuracy \\
    \midrule
    $C_1$ & Blond & 0,1 & 95.1 \\
    $C_2$ & color & 0,1 & 98.0 \\
    \bottomrule
\end{tabular}
\end{table}

\section{Discussion}
\subsection{2D Gaussian: Workings of \ourmethod{} in closed form\label{appsec:2d-gaussian}}
\begin{figure}[ht]
    \centering
    \subcaptionbox{\parbox{0.2\textwidth}{ True underlying data distribution}\label{fig:full-2d}}[0.24\textwidth]
    {\includegraphics[width=0.24\textwidth]{figures/simulation/full_data.pdf}}
    \hfill
    \subcaptionbox{\parbox{0.2\textwidth}{ Training data Orthogonal~Support}\label{fig:train-2d}}[0.24\textwidth]
    {\includegraphics[width=0.24\textwidth]{figures/simulation/train_data.pdf}}
    \hfill
    \subcaptionbox{\parbox{0.2\textwidth}{Conditional distribution learned by vanilla diffusion objective\label{fig:vanilla-2d}}}[0.24\textwidth]
    {\includegraphics[width=0.24\textwidth]{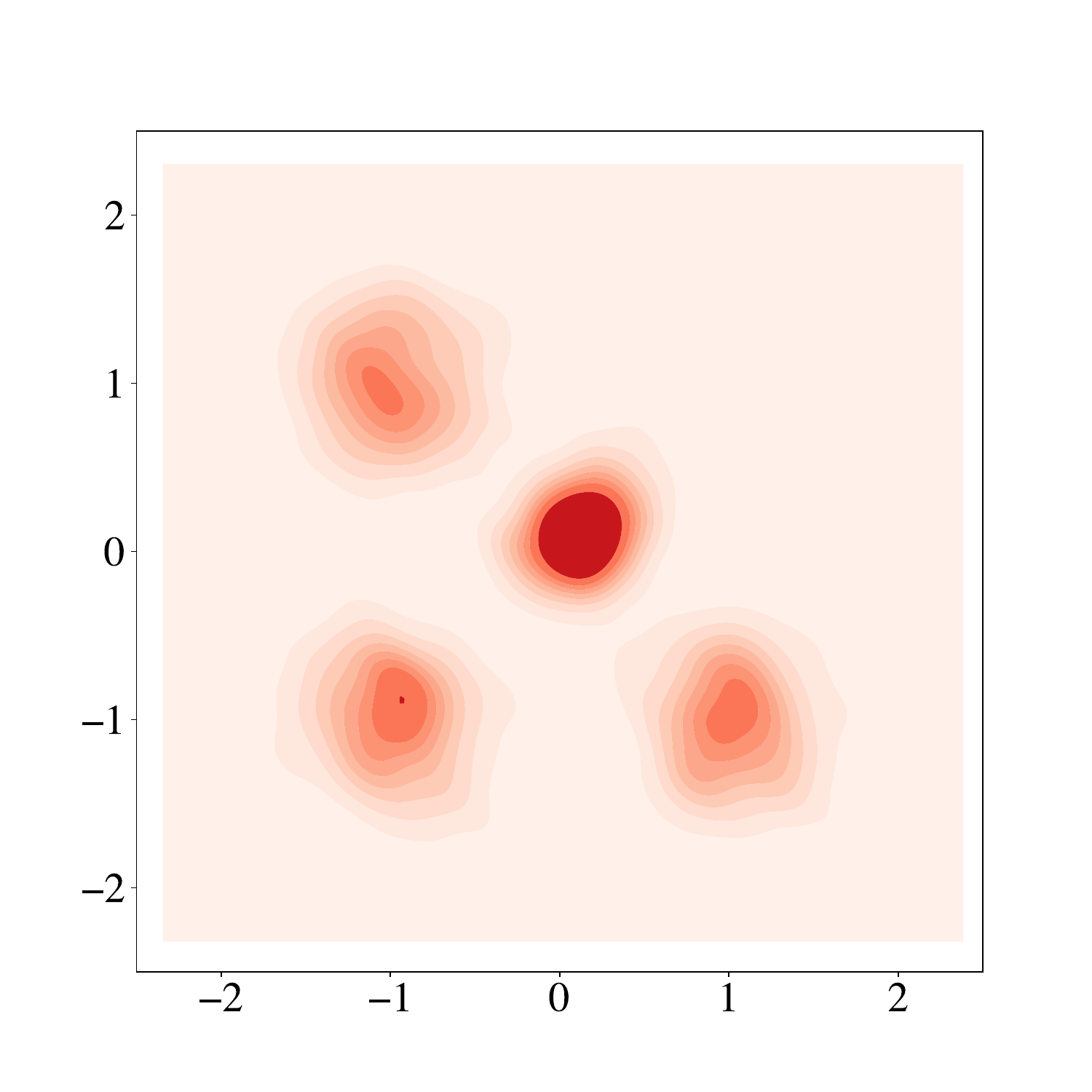}}
    \hfill
    \subcaptionbox{\parbox{0.2\textwidth}{ Conditional distribution learned by \ourmethod{}\label{fig:coind-2d}}}[0.24\textwidth]
    {\includegraphics[width=0.24\textwidth]{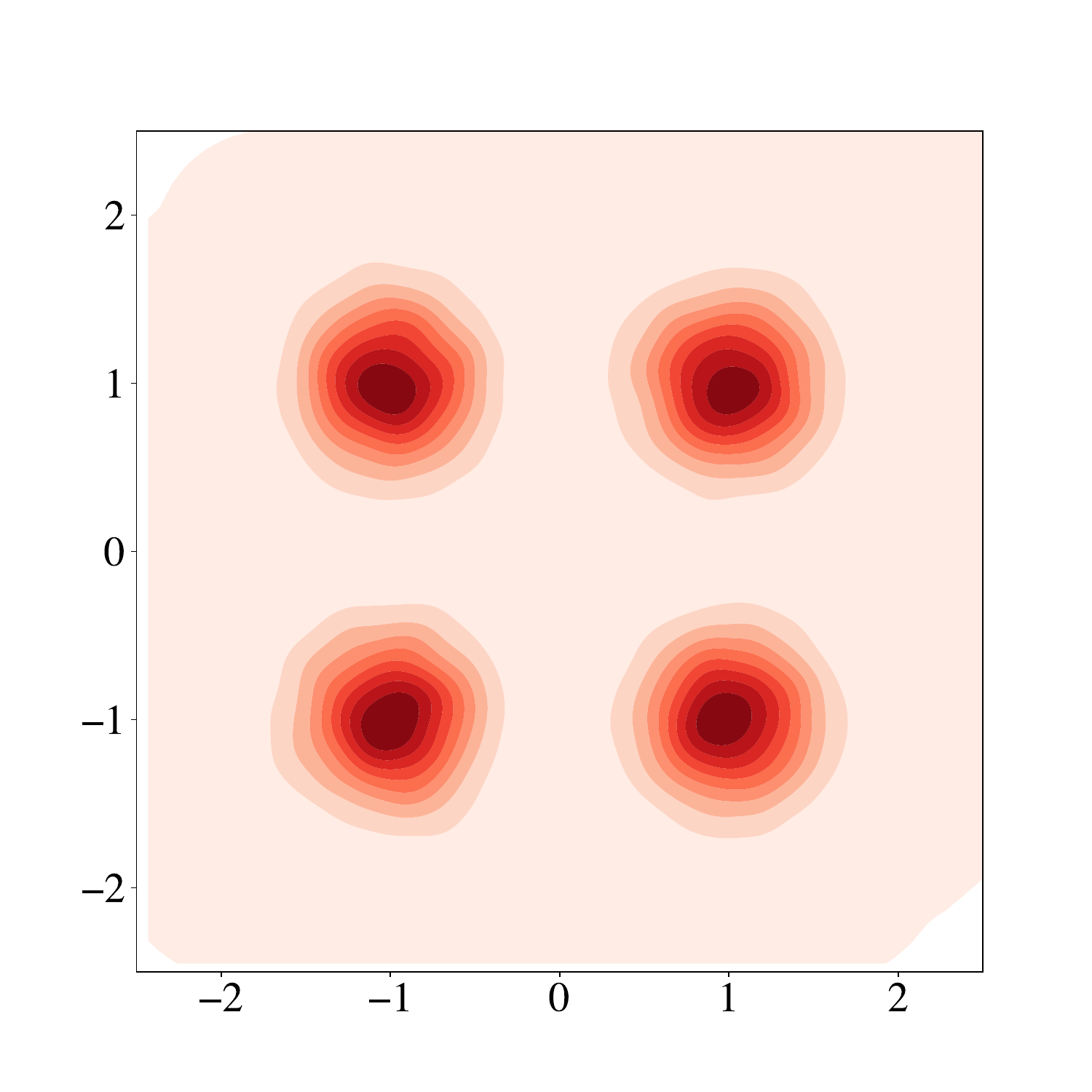}}
    \caption{\ourmethod{} respects underlying independence conditions thereby generating true data distribution (d). }
    \label{fig:2d-gaussian}
\end{figure}
The underlying data is generated by two independent attributes, $C_1$ and $C_2$. The observed variable $\mathbf{X}$ is defined as:
\begin{equation}
    \mathbf{X} = f(\mathbf{C}_1) + f(\mathbf{C}_2)\label{eq:2d-gaussian}
\end{equation}
where $f(c_i) = c_i + \sigma\epsilon$, and $\epsilon \sim \mathcal{N}(0,I)$ represents Gaussian noise. For simplicity, $C_1$ and $C_2$ are binary variables taking values in $\{-1, +1\}$.
The function $f(C_1)$ results in a mixture of Gaussians with means $\begin{bmatrix}-1 & 0\end{bmatrix} \text{ and } \begin{bmatrix}+1  & 0\end{bmatrix}$. These are represented along the x-axis in Figure \ref{fig:full-2d}. Similarly, $f(C_2)$ produces a mixture of Gaussians with means:$\begin{bmatrix}0  & -1\end{bmatrix} \text{ and } \begin{bmatrix}0  & +1\end{bmatrix}$. These are displayed along the y-axis in Figure \ref{fig:full-2d}. The combination of $C_1$ and $C_2$ independently generates as \cref{eq:2d-gaussian} This results in a two-dimensional Gaussian mixture, as illustrated in Figure \ref{fig:2d-gaussian}. We consider orthogonal support, where attribute combinations of $(C_1,C_2) \in \{(-1,-1),(-1,+1),(+1,-1)\}$, and the model is tasked to generate unseen combination of $(+1,+1)$. Also as a reminder that assumptions mentioned in \cref{sec:problem-def} are satisfied. (1) $C_1,C_2$ independently generate $\X$, and (2) all possibles values for every attribute are present at-least observed during training. Let score is given by $s_{+1,+1}$ represents $\score p(X\mid C_1=1,C_2=1)$ and likewise $s_{1,\varnothing}$ represents $\score p(X\mid C_1)$
To sample for the unseen compositions of (1,1) we use \cref{eq:and-score} to 
\begin{equation}
    s_{1,1}(x) = s_{1,\varnothing}(x) + s_{\varnothing,1}(x) - s_{\varnothing,\varnothing}(x) 
\end{equation}
Training diffusion model (score) objective involves computing score functions from the training data, which will give us the following terms in closed form. For example $s_{1,\varnothing}(x)$ is training using only $+1,-1$ combinations present during training. which is nothing but a gaussian at $+1,-1$ and the score of the gaussian is given in closed form. 
\begin{align*}
    s_{1,\varnothing}(x) &= \frac{\mu_{1,-1} - x}{\sigma^2} \\
    s_{\varnothing,1}(x) &= \frac{\mu_{1,-1} - x}{\sigma^2}
\end{align*}
$s_{\varnothing,\varnothing}(x)$ is a mixture of gaussian with means around 3 Gaussians present during training. The score of the mixture gaussian as: 
\begin{equation*}
    s_{\varnothing,\varnothing}(x) = \frac{\sum_i \mathcal{N}(x; \mu_i,\sigma^2I) \left[\frac{\mu_{i} - x}{\sigma^2}\right]}{\sum_i \mathcal{N}(x; \mu_i,\sigma^2I)}
\end{equation*}
Now leveraging Langevin dynamics \cref{og_sampling} will generate the \cref{fig:vanilla-2d} as the distribution of $P(X\mid C_1=+1,C_2=+1)$ is incorrect ( strong red blob between the $(+1,-1),(-1,+1)$ instead of gaussian at (+1,+1)). This is due to incorect modelling of the distributions $s_{1,\varnothing}(x), s_{\varnothing,1}(x), s_{\varnothing,\varnothing}(x)$. However, \ourmethod{} does not explicitly model $s_{1,\varnothing}$, instead learn joint $s_{-1,-1}(x),s_{+1,-1}(x),s_{-1,+1}(x)$ as Gaussians and then combine them using pairwise conditional independence conditions given as:
\begin{align*}
s_{-1,-1}(x) &= s_{+1,\varnothing}(x) + s_{\varnothing,+1}(x) - s_{\varnothing,\varnothing}(x) \\
s_{+1,-1}(x) &= s_{+1,\varnothing}(x) + s_{\varnothing,-1}(x) - s_{\varnothing,\varnothing}(x) \\
s_{-1,+1}(x) &= s_{-1,\varnothing}(x) + s_{\varnothing,+1}(x) - s_{\varnothing,\varnothing}(x) \\
s_{+1,1}(x) &= s_{+1,\varnothing}(x) + s_{\varnothing,+1}(x) - s_{\varnothing,\varnothing}(x) \\
&= s_{+1,-1}(x) +s_{-1,+1}(x) - s_{-1,-1}(x) \\
&= \frac{\left[\mu_{+1,-1} +\mu_{+1,-1} - \mu_{-1,-1}\right]  - x}{\sigma^2}
\end{align*}
This shows the workings of \ourmethod{} and also demonstrates that conditional independence constraints are necessary to learn the underlying distribution and alos with these constraints, diffusion models generate incorrect interpolation for unseen data distributions as shown in \cref{fig:vanilla-2d}.

\subsection{Extension to Gaussian source flow models\label{appsec:flow}}
Diffusion models can be viewed as a specific case of flow-based models where: (1) the source distribution is Gaussian, and (2) the forward process follows a predetermined noise schedule.\citep{lipman2024flowmatchingguidecode}. Can we reformulate \ourmethod{} in terms of velocity rather than score, thereby generalizing it to accommodate arbitrary source distributions and schedules?
When the source distribution is gaussian, score and velocity are related by affine transformation as detailed in Tab. 1 of \citep{lipman2024flowmatchingguidecode}. 
\begin{equation}
    s_{\theta}^t(x,C_1,C_2) = a_t x +  b_t u_{\theta}^t(x,C_1,C_2)
\end{equation}
replacing $s_{\theta}^t(\cdot)$ into \cref{eq:reduced_eq} 
\begin{align*}
    \mathcal{L}_{\text{CI}} &= \E_{p({\X,C}), t \sim U[0,1]}\E_{j,k}\lVert s_{\bm\theta}^t(x, C_j,C_k)  - s_{\bm\theta}^t(x, C_j) - s_{\bm\theta}^t(x, C_k) + s_{\bm\theta}^t(x) \rVert_2^2 \\
    &= \E_{p({\X,C}),t \sim U[0,1]} \E_{j,k} \left[ b_t^2 \lVert u_{\bm\theta}^t(x, C_j,C_k)  - s_{\bm\theta}^t(x, C_j) - u_{\bm\theta}^t(x, C_k) + u_{\bm\theta}^t(x) \rVert_2^2 \right]
\end{align*}
However we can ignore $b_t^2$, weighting for the time step $t$.
\begin{equation}\label{eq:velocity}
    \mathcal{L}_{\text{CI}} = \E_{p({\X,C}),t \sim U[0,1]} \E_{j,k} \left[ \lVert u_{\bm\theta}^t(x, C_j,C_k)  - u_{\bm\theta}^t(x, C_j) - u_{\bm\theta}^t(x, C_k) + u_{\bm\theta}^t(x) \rVert_2^2 \right]
\end{equation}
Therefore, if the source distribution is gaussian and for any arbitrary noise schedule, constraint in score translates directly to velocity constraint as given as \cref{eq:velocity}.

\end{document}